\documentclass[runningheads]{llncs}
\usepackage{graphicx}
\usepackage{hyperref}
\hypersetup{colorlinks=true,
            linkcolor=blue,
            anchorcolor=blue,
            citecolor=blue}
\usepackage{utfsym}
\usepackage{subfigure}
\usepackage{ragged2e} 
\usepackage{booktabs, makecell, multirow, tabularx}    
\begin{document}
%
\title{Noise-to-Norm Reconstruction for Industrial Anomaly Detection and Localization}
\titlerunning{Noise-to-Norm Reconstruction for Industrial Anomaly Detection}
%
\author{Shiqi Deng\inst{1,2} \and
Zhiyu Sun\inst{2} \and
Ruiyan Zhuang\inst{2}\thanks{Co-corresponding author: \email{zhuangry@midea.com}}\and
Jun Gong\inst{1}\thanks{Co-corresponding author: \email{gongjun@ise.neu.edu.cn}}}
%
%
\institute{
Northeastern University, China \and
Midea
}
%
\maketitle              
\begin{abstract}
    Anomaly detection has a wide range of applications and is especially important in industrial quality inspection. Currently, many top-performing anomaly-detection models rely on feature-embedding methods. However, these methods do not perform well on datasets with large variations in object locations. Reconstruction-based methods use reconstruction errors to detect anomalies without considering positional differences between samples. In this study, a reconstruction-based method using the noise-to-norm paradigm is proposed, which avoids the invariant reconstruction of anomalous regions. Our reconstruction network is based on M-net and incorporates multiscale fusion and residual attention modules to enable end-to-end anomaly detection and localization. Experiments demonstrate that the method is effective in reconstructing anomalous regions into normal patterns and achieving accurate anomaly detection and localization. On the MPDD and VisA datasets, our proposed method achieved more competitive results than the latest methods, and it set a new state-of-the-art standard on the MPDD dataset.
\end{abstract}
\section{Introduction}
Anomaly detection has a wide range of applications in fields such as industrial quality 
inspection~\cite{mvtec,mpdd,visa}, medical diagnosis~\cite{medicaldata}, 
and video surveillance~\cite{cuhk,shanghai}. In industrial quality inspection, 
anomaly detection can identify and locate defects in the appearance of a product, improve product quality, and ensure standards compliance. With the emergence of modern technologies in computer vision, anomaly detection using deep learning methods has rapidly developed as an effective solution for industrial quality inspections, addressing the challenges of low efficiency and difficulty in conducting large-scale manual inspections.

Supervised methods that have high data-annotation costs and low adaptability to new defects had limited use in recent years. Therefore, most studies are now focused on unsupervised learning methods. Among these, feature-embedding-based methods~\cite{padim,patchcore,fastflow,cflow} 
that utilize pretrained models to extract image features and realize the measurement or comparison of features by feature modeling are widely used. However, positional consistency of the detection objects is crucial for these methods. In contrast, reconstruction-based methods~\cite{skipgan,edg,ocrgan,mmr} 
do not have this limitation and do not require additional training data, making them suitable for various scenarios.

Reconstruction-based methods exhibit better anomaly detection and localization performances for randomly placed objects. Unlike traditional image reconstruction methods, our proposed reconstruction model uses noisy images as input; this disrupts the abnormal areas and makes it difficult to distinguish them from normal patterns, thus solving the problem of reconstructing abnormal regions owing to its strong reconstruction capabilities. In addition, our proposed reconstruction model is based on M-net~\cite{mnet} 
and employs a multiscale fusion structure. Before being fed into the reconstruction network, the noisy image is down sampled to varied sizes to enlarge the model’s receptive field, providing better robustness to anomalous regions of diverse sizes. The reconstruction network comprises three parts: an encoder, a decoder, and a feature fusion module; both the encoder and decoder contain residual attention modules and skip connections between them. The feature fusion module fuses the multiscale features to generate the reconstructed image.

Numerous experiments on the MPDD~\cite{mpdd} and VisA~\cite{visa} datasets have demonstrated that the proposed end-to-end anomaly-detection method has excellent performance. The main contributions of this study are summarized as follows:

1. We introduce a novel unsupervised anomaly detection method based on the noise-to-norm paradigm.

2. We propose a residual attention module that can be embedded in the encoder and decoder to achieve high-quality 
reconstruction of noisy images.

3. Our method achieves state-of-the-art (SOTA) performance on the MPDD dataset.

\section{Related Work}

Unsupervised learning addresses the high annotation costs and difficulty in collecting negative samples, making it the mainstream method for image anomaly detection. Unsupervised learning methods can be divided into two main categories: reconstruction- and feature embedding-based methods.

\subsection{Feature Embedding-based Methods}
Feature embedding-based methods aim to determine a feature distribution that can distinguish between normal and anomalous samples. Typically, these methods use a pre-trained network as a feature extractor to extract shallow features from images. By fitting normal sample features to a Gaussian distribution, 
Mahalanobis distance is a common method to calculates the anomaly scores~\cite{padim,fyd} between the test set samples and the Gaussian distribution, to estimate the anomaly localization.
Research in~\cite{patchcore} employed coreset-subsampled memory bank to ensure low inference cost at higher performance.
Some studies attach a normalizing flow module to the feature extractor~\cite{cflow,csflow,fastflow}, features were first extracted and a normalizing flow module was utilized to enable transformations between data distributions and well-defined densities. Subsequently, anomaly detection and localization were performed based on the probability density of the feature map.

In general, feature embedding-based methods have achieved better results on the MVTec AD~\cite{mvtec} 
dataset than those of the reconstruction-based methods because of their powerful representation capability of deep features. However, they rely on the uniformity of an object's location, which makes optimization difficult for cases in which the object's position varies significantly.

\subsection{Reconstruction-based Methods}
The reconstruction-based method trains an encoder and a decoder to reconstruct images with a low dependence on pretrained models. This method aims to train a reconstruction model that works well on positive samples but poorly on anomalous regions and achieves anomaly detection and localization by comparing the original image with the reconstructed image. Early studies used Autoencoders~\cite{dfr,edg,draem} 
for image reconstruction, whereas some methods employed a generative adversarial network~\cite{skipgan,dagan,ocrgan}
to obtain better reconstruction performance. However, there is a problem of overgeneralization, which can lead to an accurate reconstruction of anomalous regions. To address this issue, some researchers proposed a method based on image inpainting~\cite{riad,itad,maskswin,mmr}, 
in which masks are used to remove parts of the original image, preventing the reconstruction of anomalous regions. However, for images with complex structures and irregular textures, excessive loss of the original information may limit the reconstruction ability and cause many false positives in normal regions.

\section{Method}
\subsection{Overview}
\begin{figure}
    \centering
    \includegraphics[width=0.8\textwidth]{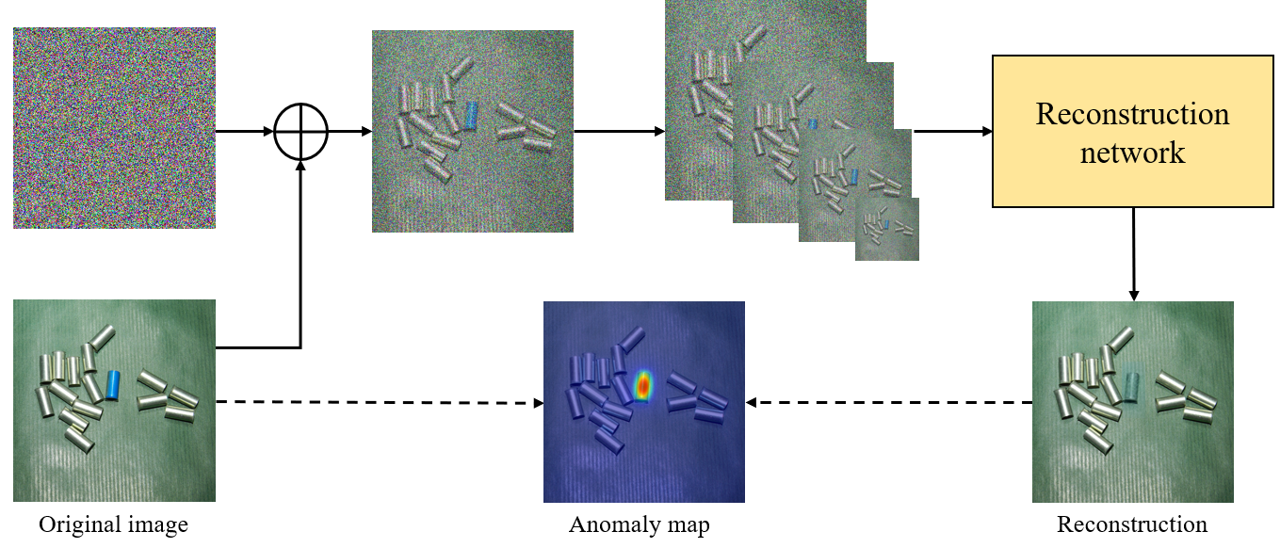}
    \caption{Framework of the proposed method.}\label{fig1}
    \end{figure}

The proposed anomaly detection framework is based on the noise-to-norm paradigm, as shown in Fig.~\ref{fig1}.

Specifically, we introduce random Gaussian noise to corrupt the original image, 
and the process of adding noise $\epsilon$ is defined as follows:
\begin{equation}
    x = (1-\lambda)x_{0} + \lambda \epsilon , \epsilon \sim{N(0.5,0.5)}
\label{eq1}
\end{equation}
where $\lambda \in(0,1),x_0 $ is the data obtained by normalizing each channel of the original image according to 
a Gaussian distribution$(\mu=0.5, \sigma=0.5)$. We add random noise generated from the same Gaussian 
distribution to the original image using weighted blending, thereby allowing us to control the 
degree to which the noise corrupts the original image. In contrast to the methods that simulate anomalies~\cite{draem}, 
our approach of adding noise is not intended to simulate anomalies. Instead, its purpose is to completely obscure the distinguishable appearance of anomalous regions, allowing the reconstruction network to transform the anomalous image into a normal image. 

After adding noise, the images were down sampled to varied sizes to serve as multiple inputs. These inputs were then utilized by the reconstruction network to generate anomaly free images. During the training phase, only anomaly free samples were used to train the reconstruction network. The reconstructed images were compared to the original images using a loss function, and the reconstruction capability of the model was continuously improved. During the inference phase, anomaly localization was achieved by generating an anomaly map that captured pixel-level differences between the reconstructed and original images. The specific details of the reconstructed network are described below.

\subsection{Reconstruction Network}
The overall architecture of the proposed reconstruction network is shown in Fig. 2. The network is based on the M-net~\cite{mnet}, 
which originated from the field of image segmentation and has been proven to be effective in the domain of denoising. Inspired by the SRMnet~\cite{srmnet}, 
we incorporated pixel shuffle operations into the encoder and decoder for upsampling and downsampling; this allows us to effectively manage resolution changes in the network and improve the reconstruction quality. The residual attention modules were merged after concatenating the features to enhance the feature representation and capture the relevant information. The encoder and decoder were connected through skip connections to facilitate the flow of information between different feature levels. The multiscale features were combined in the feature fusion module to generate the final reconstructed image. This design enables the network to effectively capture anomalies and produce high-quality reconstructions.

\begin{figure}
    \centering
    \includegraphics[width=0.9\textwidth]{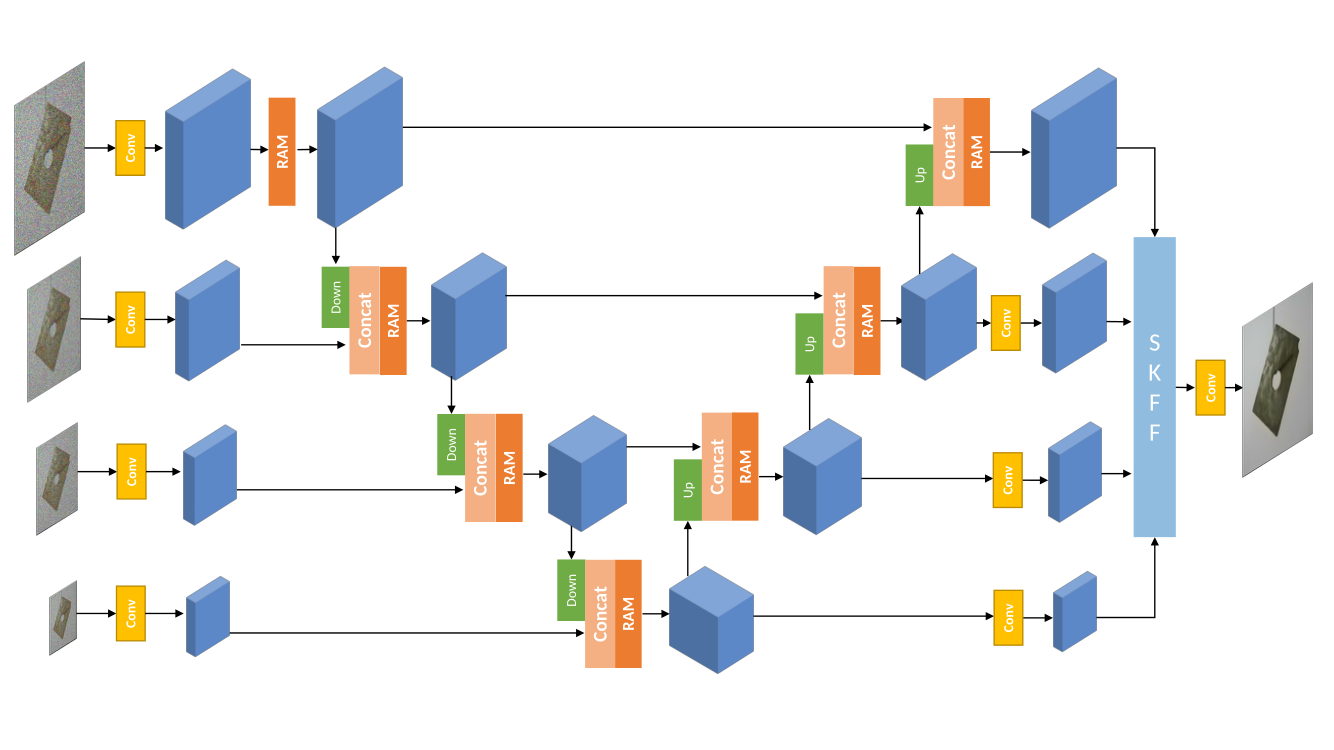}
    \caption{The overall architecture of the proposed reconstruction network.}\label{fig2}
    \end{figure}

\subsubsection{Residual Attention Module} 
\begin{figure}
    \centering
    \includegraphics[width=0.7\textwidth]{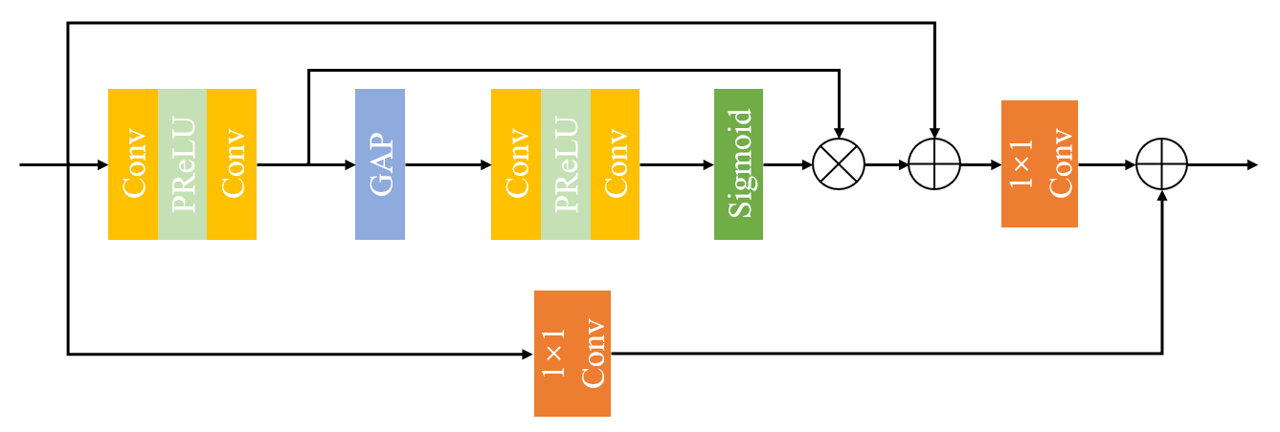}
    \caption{Illustration of the residual attention module.}\label{fig3}
    \end{figure}

The Residual Attention Modules are integrated after the concatenation of features to enhance feature representation and capture relevant information. These modules leverage residual connections and attention mechanisms to selectively emphasize notable features and suppress irrelevant ones. By focusing on informative regions and enhancing feature discrimination, the residual attention modules improve the network’s ability to generate high-quality reconstructions. In addition, the residual connections address the issue of vanishing gradients. By propagating gradients more effectively through the network, the residual connections enable faster convergence and improve the accuracy of the model. The specific structure of the residual attention module is shown in Fig.~\ref{fig3}. 
It comprises global pooling, convolutional, and activation layers. In both pathways, a $1\times1$ 
convolutional layer is employed to adjust the number of feature channels.

\subsubsection{Selective Kernel Feature Fusion (SKFF)} 
Our decoder generates four feature maps with different resolutions, and we employed the SKFF~\cite{skff} 
module for feature fusion. The SKFF allows for the selection of different convolutional kernels at different spatial positions to facilitate the fusion of features from different scales, enabling the integration of multiscale reconstruction features. This approach avoids directly connecting each feature map and instead aggregates weighted features, addressing the issues of a large number of parameters and higher computational complexity in the M-net. 




\subsection{Metric Function} 
We employed a metric function that combines the MS-SSIM and $\ell_1$ proposed by Hang Zhao et al.in~\cite{msssim}. SSIM~\cite{ssim} is a 
widely used indicator for measuring the structural similarity between images. The SSIM for pixel $p$ is 
defined in Eq.~\ref{2}.
\begin{equation}
SSIM(p) = \frac{2\mu_{x}\mu_{y}+C_{1}}{\mu_{x}^2+\mu_{y}^2+C_{1}} \cdot \frac{2\sigma_{xy}+C_{2}}{\sigma_{x}^2+\sigma_{y}^2+C_{2}}
        = l(p)\cdot cs(p)
        \label{2}
\end{equation}
where the means 
and standard deviations are computed using a Gaussian filter $G_{\sigma_G}$ with a standard deviation $\sigma_G$. 

MS-SSIM uses different Gaussian filters ($\sigma ={0.5, 1, 2, 4, and 8}$) to compute the original image and is, defined as follows:

\begin{equation}
    MS\mbox{-}SSIM(p) = l_{M}^{\alpha}(p) \cdot \prod_{j=1}^{M} cs_{j}^{\beta_{j}}(p)
        \label{3}
\end{equation}

where $l_{M}$ and $cs_{j}$ are the terms defined in Eq.~\ref{2}, and the index $j$ represents different Gaussian 
filters with different $\sigma$ values, For convenience, we set $\alpha = \beta_{j} =1$, for $j = \left\{1,\ldots,M\right\}$. 

During the training phase, 
for an image of size $H \times W$, the MS-SSIM loss can be expressed as follows:
\begin{equation}
    \mathcal{L}_{MS\mbox{-}SSIM} = \frac{1}{H \times W} \sum_{p} 1-MS\mbox{-}SSIM(p)
        \label{4}
\end{equation}

The total loss is calculated by adding the 
$\ell_1$ loss multiplied by the Gaussian filter and the weighted MS-SSIM loss. This formula is shown 
as Eq.~\ref{5}.
\begin{equation}
    \mathcal{L}_{total} = \alpha \cdot \mathcal{L}_{MS\mbox{-}SSIM} + (1-\alpha) \cdot G_{\sigma_G^M} \cdot \mathcal{L}_{l_1}
        \label{5}
\end{equation}
where $\alpha$ represents the weight coefficient.

During the inference stage, 
we calculated the anomaly localization by computing the MS-SSIM and  $\ell_1$ error for each pixel.

\section{Experiments}

\subsection{Datasets} 

\subsubsection{MPDD} 
MPDD~\cite{mpdd} is a challenging dataset that focuses on detecting defects in the manufacturing process of painted metal parts. It reflects the real-world situations encountered by human workers on production lines. The dataset includes six categories of metal parts. The images were captured under various spatial orientations, positions, and distance conditions with different light intensities and non-uniform backgrounds. The training set consisted of 888 normal samples, whereas the test set consisted of 176 normal and 282 abnormal samples.

\subsubsection{VisA} 
VisA~\cite{visa} consists of 10,821 images. There are 9,621 normal and 1,200 abnormal images. 
VisA contains 12 subsets, each corresponding to one class of objects. 
We assigned 90$\%$ of the normal images to the training set, whereas
10$\%$ of the normal images and all anomalous samples were grouped as the test set.

\subsection{Experimental Details} 
OOur studywork wais implemented in PyTtorch usingwith an NVIDIA GeForce GTX 2080Ti. We resized all the original images of the VisA and the MPDD images to $256 \times 256$ 
for both training and testing. We divided 20$\%$ of the training dataset 
into validation sets. For each category of these two datasets, we utilized AdamW optimizer~\cite{adamw} 
with $\beta = (0.5, 0.999)$. We set the initial learning rate to $10^{-6}$ and used cosine annealing~\cite{cos} to adjust the 
learning rate with $T\_\max=100$ and $eta\_\min=10^{-6}$. The maximum number of training epochs was set to 500, and the
training was stopped early if the loss did not decrease within 20 consecutive epochs.

We evaluated our approach using different metrics for comparison with other baselines. We used the area under the curve (AUC) of the receiver operating characteristic (ROC) to evaluate the performance of image-level anomaly detection and pixel-level anomaly localization.

\subsection{Comparative Experiments} 
\subsubsection{MPDD} 
\begin{figure}[htb]
    \centering
    \subfigure[]{
        \begin{minipage}[b]{0.1\linewidth}
            \centering
            \includegraphics[width=1\linewidth]{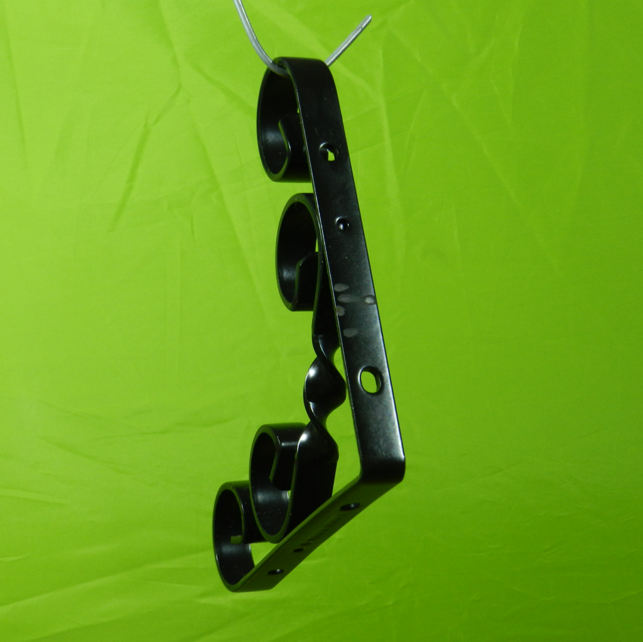}
            \includegraphics[width=1\linewidth]{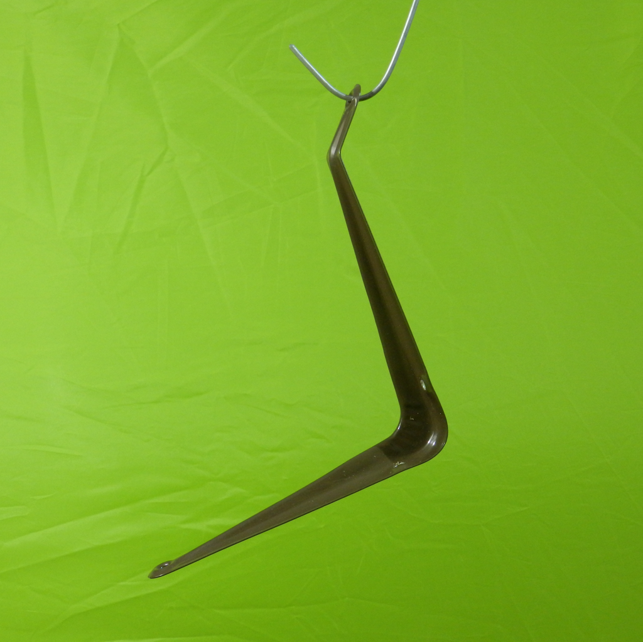}
            \includegraphics[width=1\linewidth]{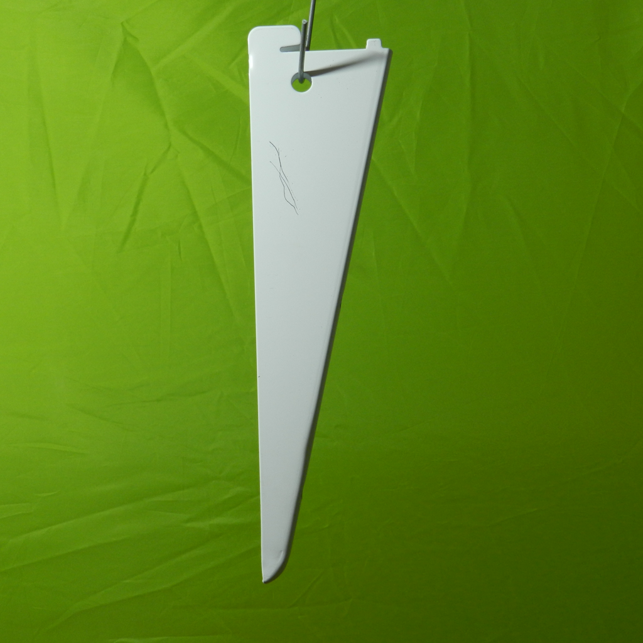}
            \includegraphics[width=1\linewidth]{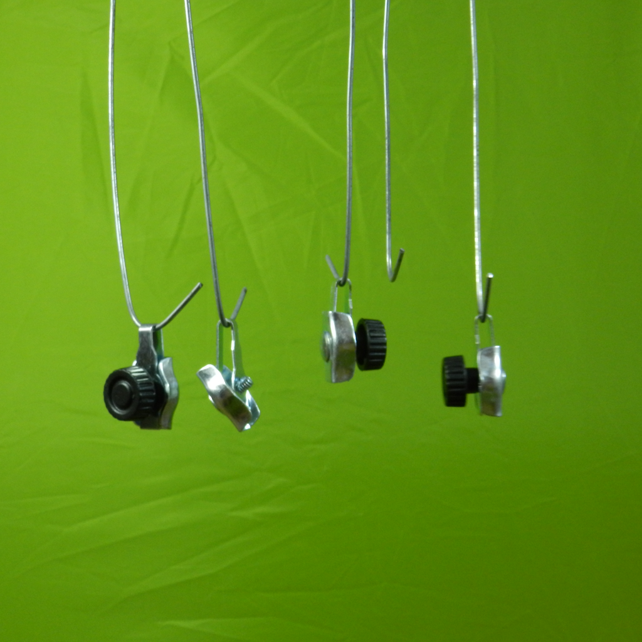}
            \includegraphics[width=1\linewidth]{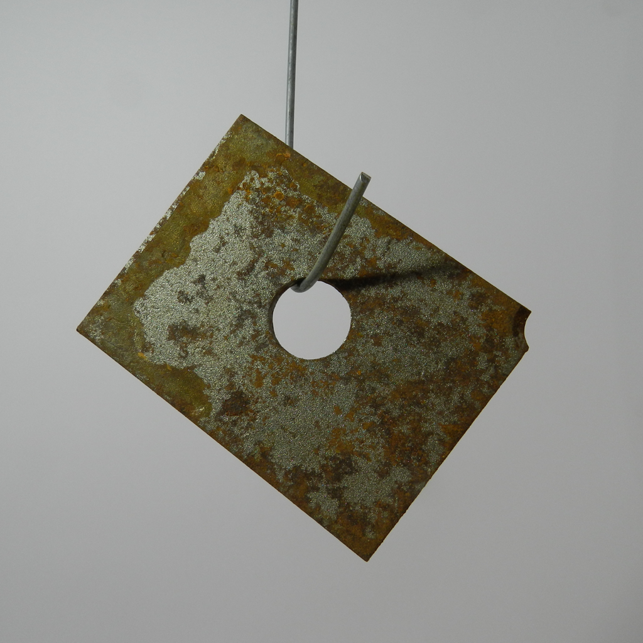}
            \includegraphics[width=1\linewidth]{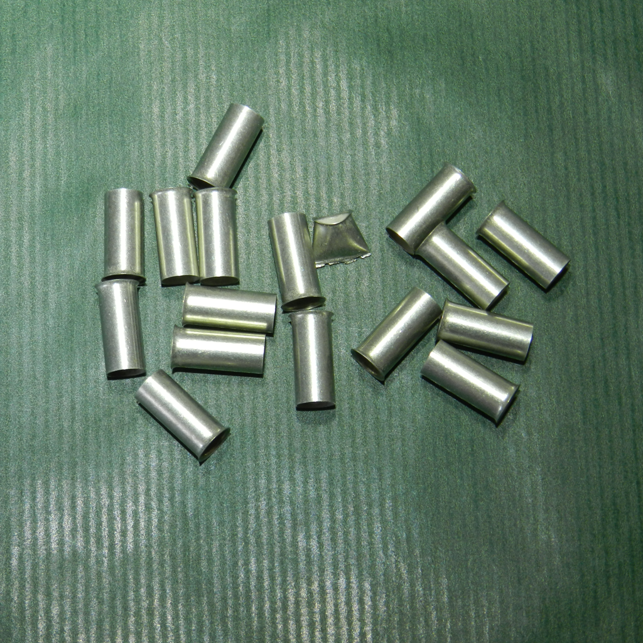}
        \end{minipage}}\hspace{-1mm}
    \subfigure[]{
         \begin{minipage}[b]{0.1\linewidth}
            \centering
            \includegraphics[width=1\linewidth]{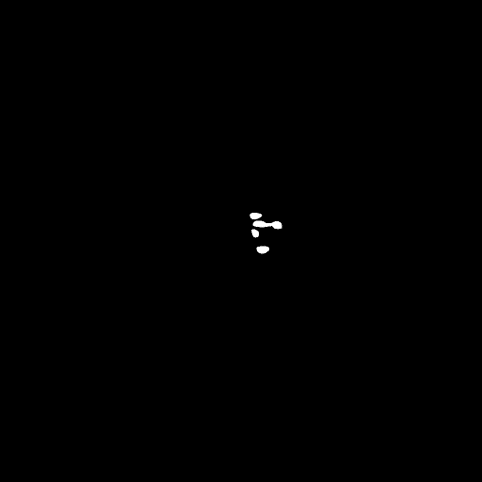}
            \includegraphics[width=1\linewidth]{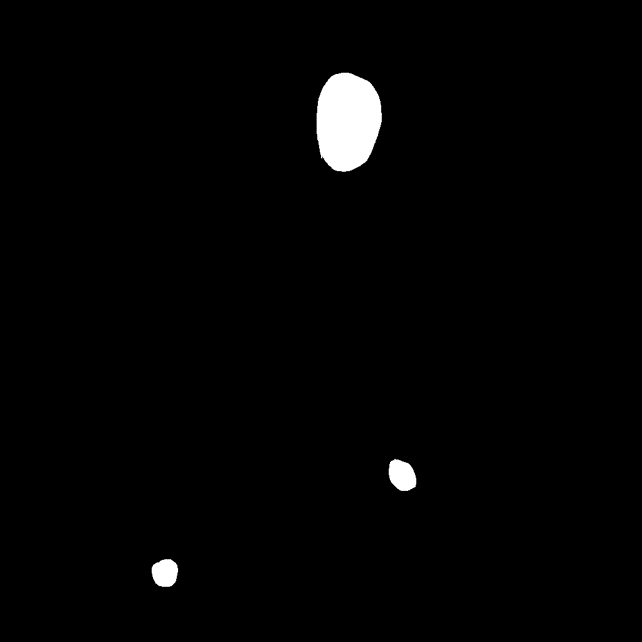}
            \includegraphics[width=1\linewidth]{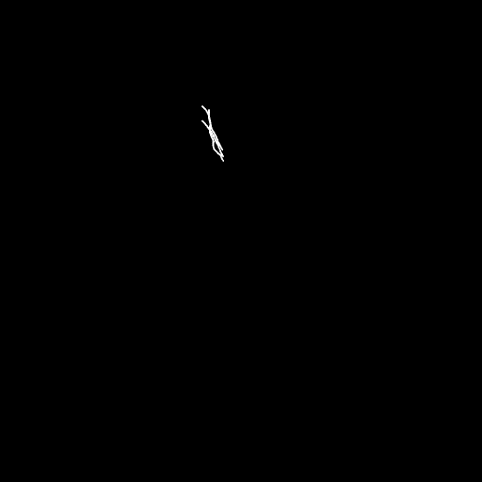}
            \includegraphics[width=1\linewidth]{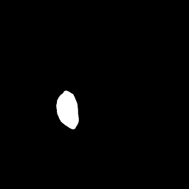}
            \includegraphics[width=1\linewidth]{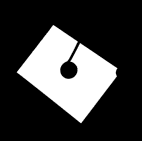}
            \includegraphics[width=1\linewidth]{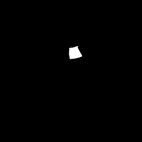}
        \end{minipage}}\hspace{-1mm}
    \subfigure[]{
         \begin{minipage}[b]{0.1\linewidth}
            \centering
            \includegraphics[width=1\linewidth]{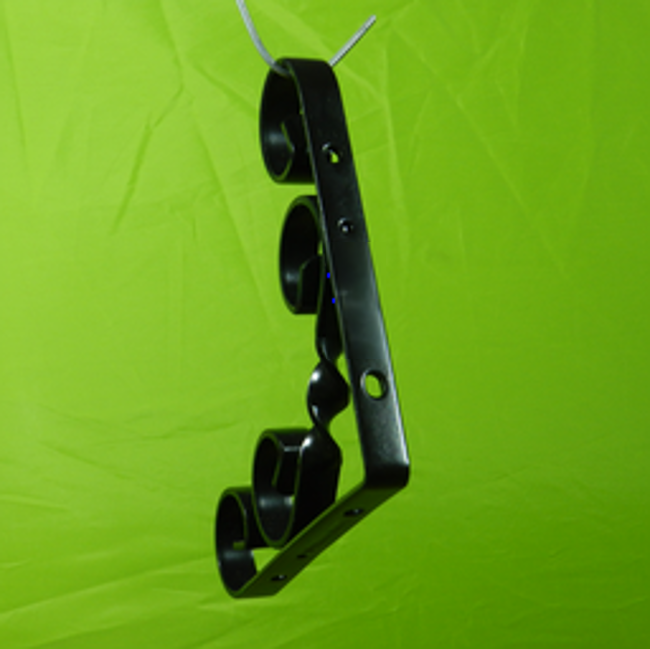}
            \includegraphics[width=1\linewidth]{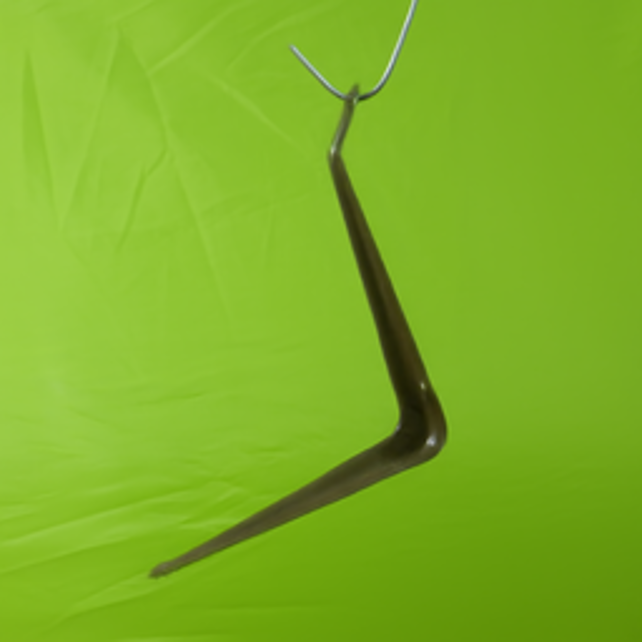}
            \includegraphics[width=1\linewidth]{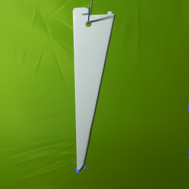}
            \includegraphics[width=1\linewidth]{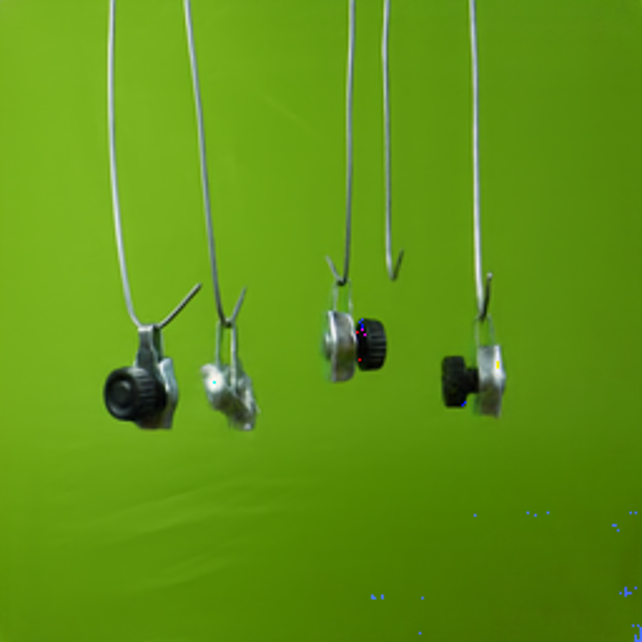}
            \includegraphics[width=1\linewidth]{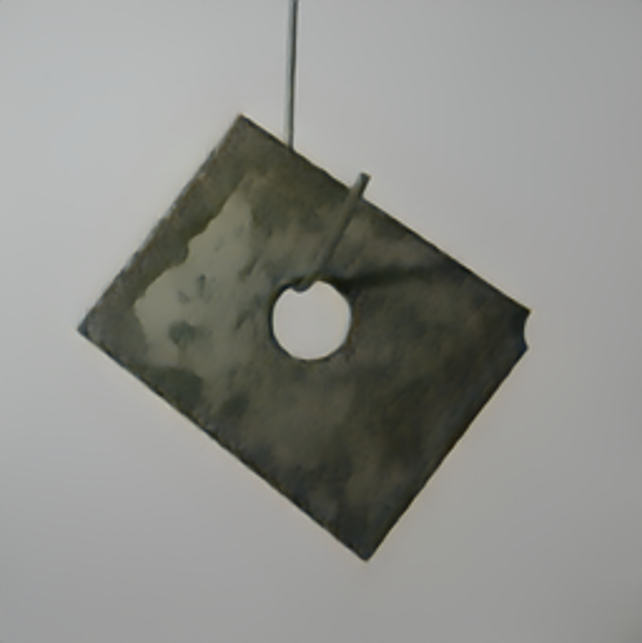}
            \includegraphics[width=1\linewidth]{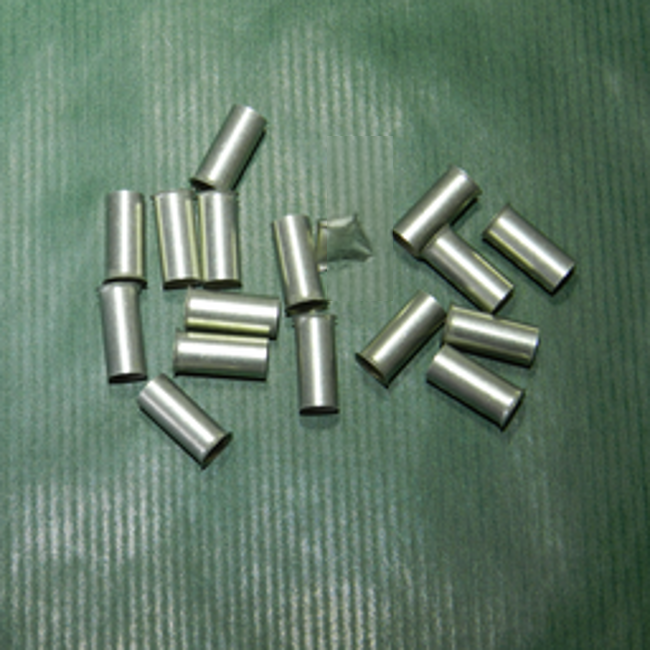}
        \end{minipage}}\hspace{-1mm}
    \subfigure[]{
        \begin{minipage}[b]{0.1\linewidth}
            \centering
            \includegraphics[width=1\linewidth]{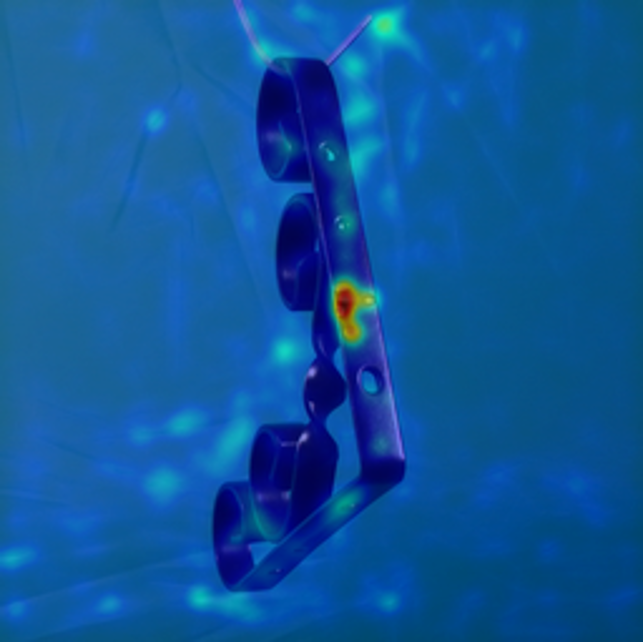}
            \includegraphics[width=1\linewidth]{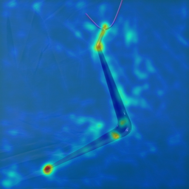}
            \includegraphics[width=1\linewidth]{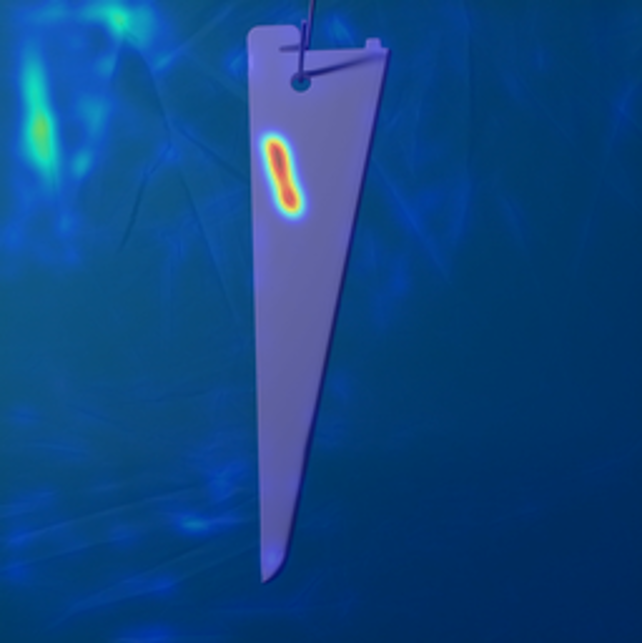}
            \includegraphics[width=1\linewidth]{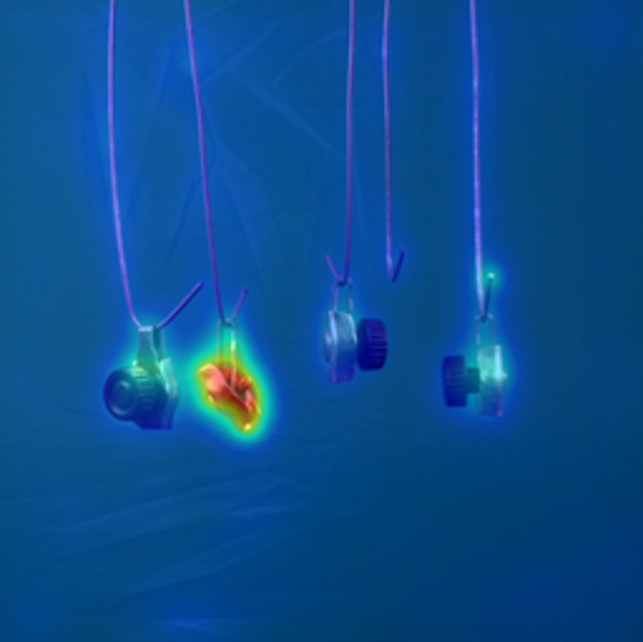}
            \includegraphics[width=1\linewidth]{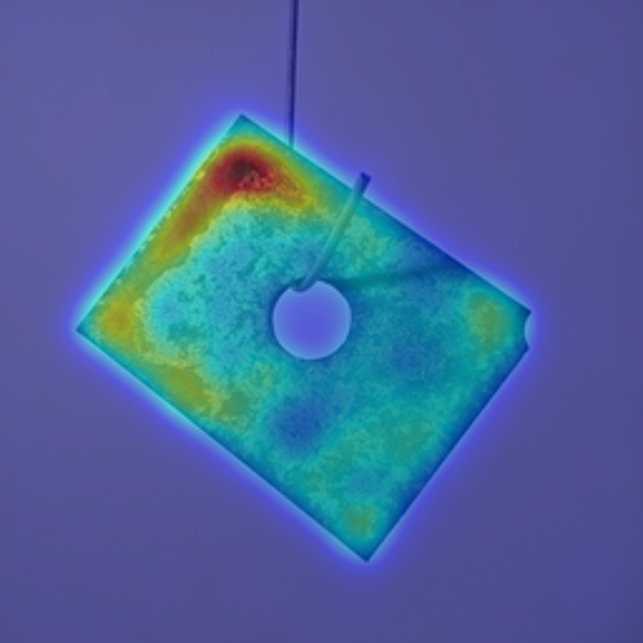}
            \includegraphics[width=1\linewidth]{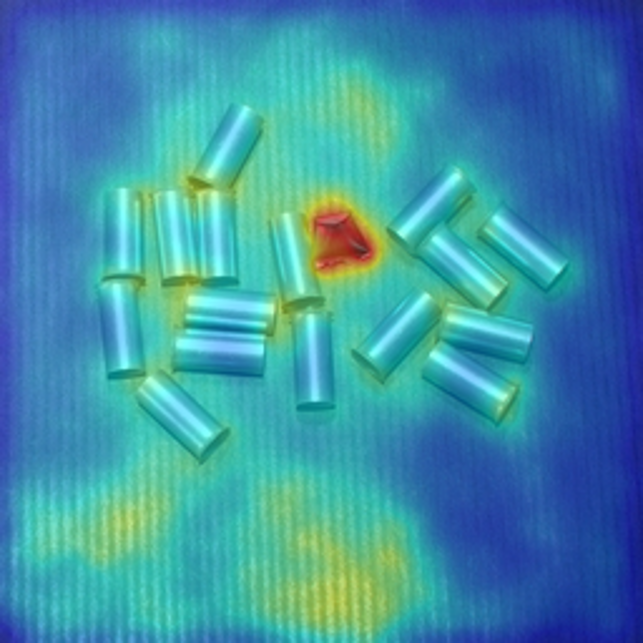}
        \end{minipage}}\hspace{-1mm}
    \subfigure[]{
         \begin{minipage}[b]{0.1\linewidth}
            \centering
            \includegraphics[width=1\linewidth]{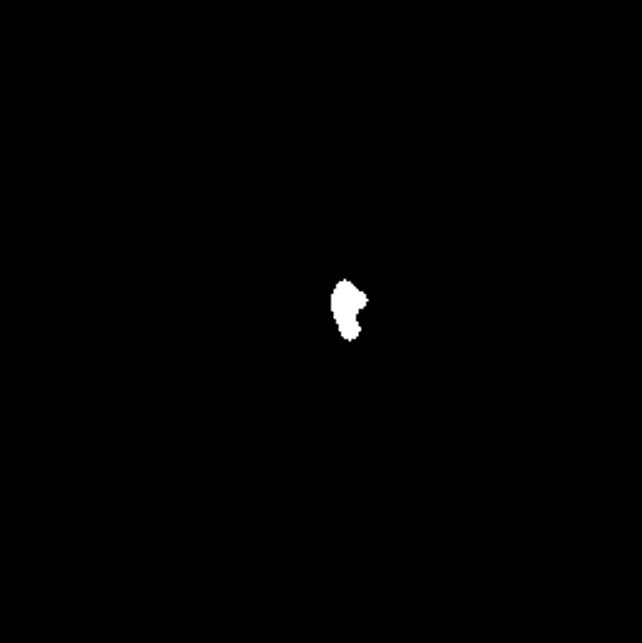}
            \includegraphics[width=1\linewidth]{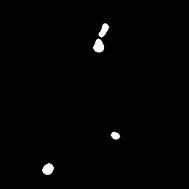}
            \includegraphics[width=1\linewidth]{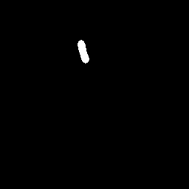}
            \includegraphics[width=1\linewidth]{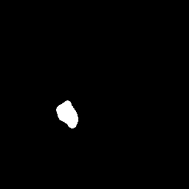}
            \includegraphics[width=1\linewidth]{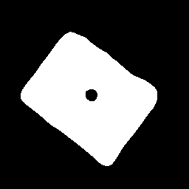}
            \includegraphics[width=1\linewidth]{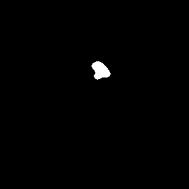}
        \end{minipage}}\hspace{-1mm}
    \subfigure[]{
         \begin{minipage}[b]{0.1\linewidth}
            \centering
            \includegraphics[width=1\linewidth]{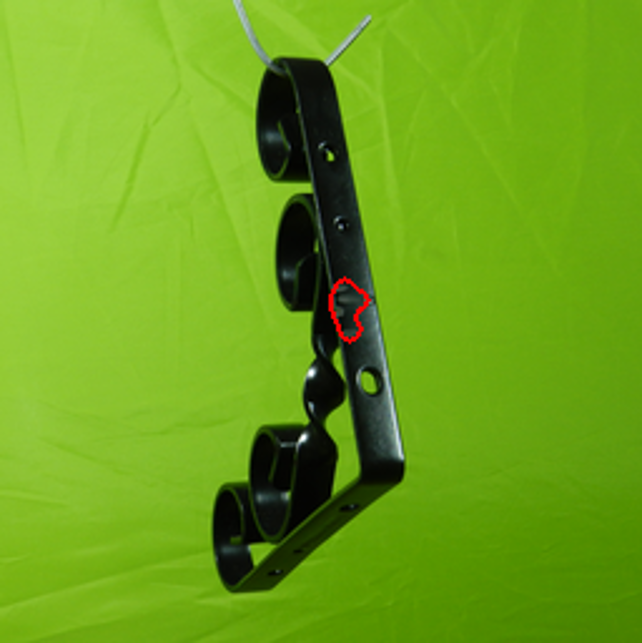}
            \includegraphics[width=1\linewidth]{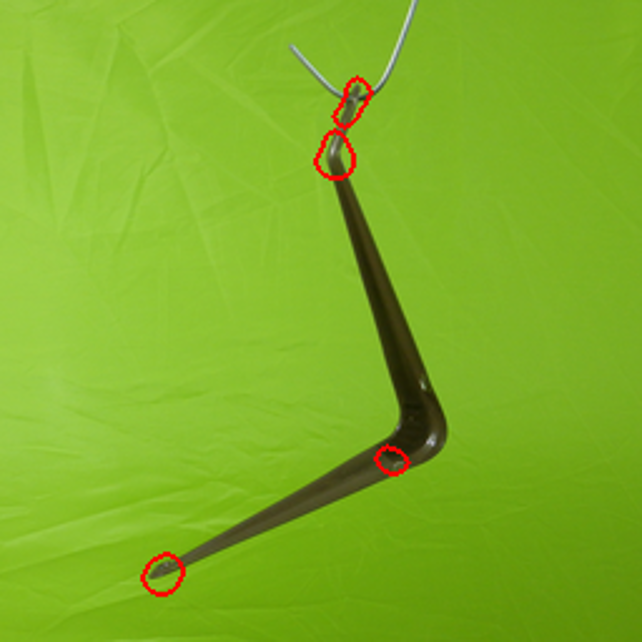}
            \includegraphics[width=1\linewidth]{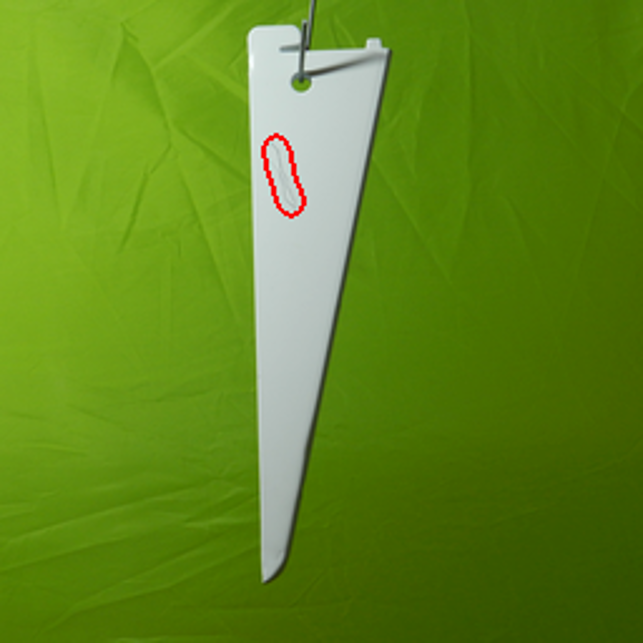}
            \includegraphics[width=1\linewidth]{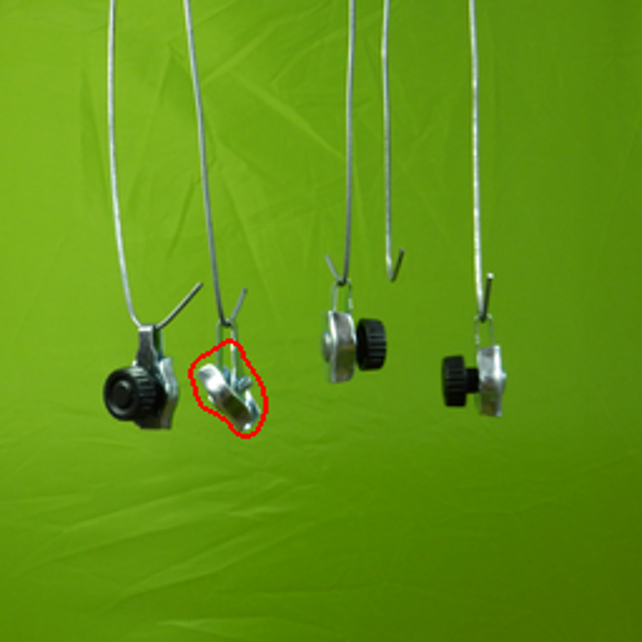}
            \includegraphics[width=1\linewidth]{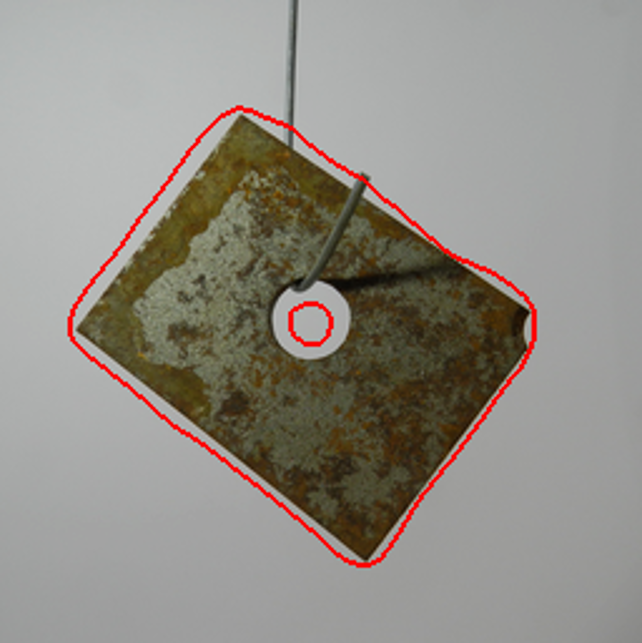}
            \includegraphics[width=1\linewidth]{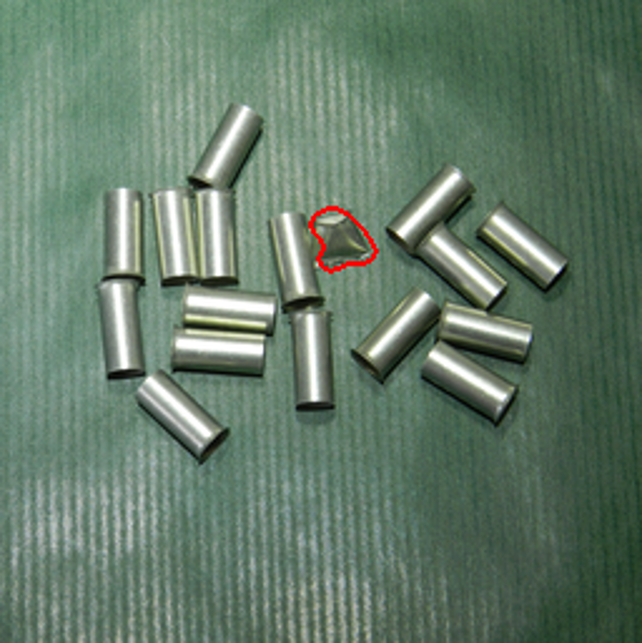}
        \end{minipage}}\hspace{-1mm}
    \caption{Visualization of examples on MPDD\@. (a) Original image (b) Ground truth (c) Reconstructed image 
    (d) Anomaly map (e) Prediction of anomalous regions (f) Prediction of anomalous regions on the original image}\label{fig4}
    \end{figure}

We compared theour proposed method with several state-of-the-artSOTA methods on the MPDD dataset, including reconstruction-based methods~\cite{dagan,skipgan} 
and feature-based methods~\cite{padim,cflow,patchcore}.
The image-level detection results are listed in Table~\ref{tab1}, and the anomaly 
segmentation results are presented in Table~\ref{tab2}. Experiments demonstrated that our proposed method outperformed 
previous SOTA methods on the MPDD dataset.
The partial visualization results of the proposed method on the MPDD~\cite{mpdd} dataset are shown in Fig.~\ref{fig4}.

\begin{table}
    \centering
    \caption{Comparison of image-level detection results (AUROC$\%$) on the MPDD dataset.Best results are highlighted in bold.
    }\label{tab1}
    \setlength{\tabcolsep}{1.2mm}{
        \begin{tabular}{lcccccc}
        \hline
        Method       & DAGAN & Skip-GANomaly & PaDiM & CFLOW & PatchCore & Ours\\
        \hline
        Bracket Black & 68.55 & 61.30 & 75.60 & 72.67 & 81.88              & \textbf{93.42}\\
        Bracket Brown & 77.07 & 62.14 & 85.40 & 88.84 & 78.43              & \textbf{93.14}\\
        Bracket White & 72.11 & 73.33 & 82.22 & 87.78 & 76.00              & \textbf{89.33}\\
        Connector     & 99.76 & 73.62 & 91.67 & 94.76 & 96.67              & \textbf{100.00}\\
        Metal Plate   & 85.43 & 73.24 & 56.30 & 99.51 & \textbf{100.00}    & 99.57\\
        Tubes         & 31.93 & 46.42 & 57.51 & 73.14 & 59.73              & \textbf{94.16}\\
        \hline
        Avg.          & 72.48 & 64.84 & 74.78 & 86.12 & 82.12              & \textbf{94.94}\\
        \hline
        \end{tabular}}
    \end{table}

\begin{table}
    \centering
    \caption{Comparison of pixel-level detection results (AUROC$\%$) on the MPDD dataset.Best results are highlighted in bold.
    }\label{tab2}
    \setlength{\tabcolsep}{1.2mm}{
        \begin{tabular}{lcccccc}
        \hline
        Method        & DAGAN & Skip-GANomaly & PaDiM & CFLOW  &PatchCore& Ours\\
        \hline
        Bracket Black & 89.73 & 88.96 & 94.23 & 96.88 & 98.41            & \textbf{98.97}\\
        Bracket Brown & 81.50 & 78.07 & 92.44 & \textbf{97.78} & 91.46   & 93.10\\
        Bracket White & 70.63 & 78.81 & 98.11 & \textbf{98.61} & 97.44   & 97.82\\
        Connector     & 85.73 & 80.20 & 97.89 & 98.39 & 95.00            & \textbf{98.95}\\
        Metal Plate   & 89.95 & 89.72 & 92.93 & 98.21 & 96.57            & \textbf{98.78}\\
        Tubes         & 82.31 & 77.30 & 93.94 & 96.43 & 95.05            & \textbf{99.17}\\
        \hline
        Avg.          & 83.31 & 82.19 & 96.74 & 97.72 & 95.66            & \textbf{97.80}\\
        \hline
        \end{tabular}}
    \end{table}

Specifically, as shown in Table~\ref{tab1},  our proposed method achieved an overall improvement of 8.82$\%$ 
compared to that of the previous best-performing method, CFLOW~\cite{cflow}. 
The most significant improvement was observed in the tubes that contained multiple instances with a random distribution of positions. These results highlighted the advantages of the proposed method.
As shown in Table~\ref{tab2}, the best average performance was achieved. 
However, the proposed method has some limitations. We were unable to achieve satisfactory performance in the brown bracket category. Most defects in the brown bracket category are deformation defects, and our method cannot accurately restore deformations, which hinders the accurate identification of such defects.

\subsubsection{VisA} 
Further, to validate the generalizability and versatility of our method, we compared it with other SOTA methods~\cite{draem,rd4ad,padim,cflow,fastflow,patchcore} 
on the VisA~\cite{visa} dataset. 
The anomaly detection results for the VisA dataset are listed in Table~\ref{tab3}. 
Experiments demonstrated that our proposed method performed competitively on the VisA dataset.

\begin{table}
    \centering
    \caption{Comparison of image-level detection results (AUROC$\%$) on the VisA dataset.Best results are highlighted in bold.
    }\label{tab3}
    \setlength{\tabcolsep}{1.2mm}{
    \begin{tabular}{lccccccc}
    \hline
    Method   & DRAEM & RD4AD & PaDiM & CFLOW & FastFlow & PatchCore & Ours\\
    \hline
    Candle      & 94.4 &92.2&91.6&97.0&92.8&\textbf{98.6}&83.7 \\
    Capsules    & 76.3 &90.1&70.7&93.0&71.2&81.6&\textbf{93.3} \\
    Cashew      & 90.7 &\textbf{99.6}&93.0&90.9&91.0&97.3&93.4 \\
    Chewing gum & 94.2 &\textbf{99.7}&98.8&98.3&91.4&99.1&97.7 \\
    Fryum       & 97.4 &96.6&88.6&91.1&88.6&96.2&\textbf{97.3} \\
    Macaroni1   & 95.0 &\textbf{98.4}&87.0&69.6&98.3&97.5&91.6 \\
    Macaroni2   & 96.2 &\textbf{97.6}&70.5&77.2&86.3&78.1&91.5 \\
    PCB1        & 54.8 &97.6&94.7&91.4&77.4&\textbf{98.5}&94.7 \\
    PCB2        & 77.8 &91.1&88.5&96.7&61.9&\textbf{97.3}&95.6 \\
    PCB3        & 94.5 &95.5&91.0&\textbf{99.6}&74.3&97.9&98.7 \\
    PCB4        & 93.4 &96.5&97.5&94.2&80.9&\textbf{99.6}&98.2 \\
    Pipe fryum  & 99.4 &97.0&97.0&99.0&72.0&\textbf{99.8}&92.6 \\
    \hline
    Avg.        & 88.7 &\textbf{96.0}&89.1&91.5&82.2&95.1&94.0 \\
    \hline
    \end{tabular}}
    \end{table}

\subsection{Ablation Studies} 
\subsubsection{Effect of $\lambda$} 
In this study, we employed a noise-to-norm reconstruction paradigm. To validate the effectiveness of adding noise and the effect of the noise coefficient ($\lambda$) 
oon the detection results, we conducted comparative experiments. The results, as shown in Table~\ref{tab4}, 
demonstrate that the overall detection performance was the best when $\lambda=0.3$. Compared to that of the case without added noise ($\lambda=0$), 
the detection accuracy increased by $22.28\%$, and the segmentation accuracy increased by $9.55\%$. Therefore, 
we finally set $\lambda=0.3$. These experimental results confirm the significant improvement in anomaly detection achieved using the noise-to-norm reconstruction approach.

\begin{table}
    \centering
    \caption{Effect of noise coefficient ($\lambda$) on image/pixel-level detection results (AUROC$\%$).Best results are highlighted in bold.
    }\label{tab4}
    \setlength{\tabcolsep}{2mm}{
    \begin{tabular}{lcccc}
    \hline
    Noise Coefficient  & $\lambda=0$ & $\lambda=0.2$ & $\lambda=0.3$ & $\lambda=0.4$ \\
    \hline
    Bracket Black & 47.61/77.21  & 83.71/98.16                   & \textbf{93.42}/98.97 & 90.56/\textbf{99.02}  \\
    Bracket Brown & 82.50/83.93  & 90.57/91.84                   & \textbf{93.14}/\textbf{93.10} & 84.09/92.86  \\
    Bracket White & 74.89/89.82  & \textbf{92.78}/\textbf{98.35} & 89.33/97.82 & 83.22/96.74  \\
    Connector     & 96.67/93.70  & \textbf{100.00}/98.78         & \textbf{100.00}/\textbf{98.95} & \textbf{100.00}/98.80  \\
    Metal Plate   & 97.83/97.88  & 98.92/\textbf{98.96}          & \textbf{99.57}/98.78 & 99.30/98.40  \\
    Tubes         & 39.45/85.88  & \textbf{97.06}/98.61          & 94.16/99.17 & 93.75/\textbf{99.44}  \\
    \hline
    Avg.          & 73.16/88.07  & 93.84/97.45                   & \textbf{94.94}/\textbf{97.80} & 91.82/97.54  \\ 
    \hline
    \end{tabular}}
    \end{table}

\subsubsection{Importance of Residual Attention Module} 
To demonstrate the effectiveness of the proposed residual attention module, we conducted an ablation experiment. In the control group, we replaced the residual attention module with a 
$1\times1$ convolutional layer, which was used to change the number of feature channels. The experimental results, 
as listed in Table~\ref{tab5}, indicate that adding the residual attention module improved the detection accuracy by $28.68\%$ and the segmentation accuracy by $8.94\%$. 
This demonstrates the significance of incorporating the residual attention module into the model.

\begin{table}
    \centering
    \caption{Effect of residual attention module on image/pixel-level detection results (AUROC$\%$).Best results are highlighted in bold.
    }\label{tab5}
    \setlength{\tabcolsep}{4mm}{
    \begin{tabular}{lcccc}
    \hline   
    \multirow{2}*{ Residual Attention module} & \multicolumn{2}{c}{Image-level} & \multicolumn{2}{c}{Pixel-level} \\ 
    \cmidrule(lr){2-3}\cmidrule(lr){4-5}
     & \usym{1F5F8} & \usym{2715} & \usym{1F5F8} &\usym{2715} \\
    \hline
    Bracket Black & \textbf{93.42} & 50.47 & \textbf{98.97} & 89.36  \\
    Bracket Brown & \textbf{93.14} & 75.32 & \textbf{93.10} & 80.36  \\
    Bracket White & \textbf{92.33} & 65.92 & \textbf{96.75} & 85.74  \\
    Connector     & \textbf{100.00}& 62.38 & \textbf{98.95} & 94.99  \\
    Metal Plate   & \textbf{99.57} & 88.61 & \textbf{98.78} & 96.69  \\
    Tubes         & \textbf{94.16} & 57.87 & \textbf{99.17} & 84.93  \\
    \hline
    Avg.          & \textbf{95.44} & 66.76 & \textbf{97.62} & 88.68  \\
    \hline
    \end{tabular}}
    \end{table}

\section{Conclusion}
In this study, an industrial image anomaly detection method based on noise-to-norm reconstruction is proposed. We enhanced the M-net by incorporating a residual attention module and feature fusion, obtaining a reconstruction network. Experimental results demonstrate that our method achieves SOTA performance in anomaly detection and localization on the MPDD dataset, and it also exhibits competitive performance on the VisA dataset. Our proposed method has significant advantages for handling data with multiple instances and varying object positions. However, the proposed method has limitations in detecting object absences or displacement anomalies. In future work, we will explore methods that combine the feature distribution of positive samples with a reconstruction approach to improve the anomaly detection performance of the model.

\bibliographystyle{splncs04}
\bibliography{references}
%


\end{document}